\documentclass{article}

\usepackage{cite}
\usepackage{amsmath,amssymb,amsfonts}
\usepackage{algorithmic}
\usepackage{graphicx}
\usepackage{textcomp}
\usepackage{xcolor}
\usepackage{caption,subcaption}
\newcommand\blfootnote[1]{%
  \begingroup
  \renewcommand\thefootnote{}\footnote{#1}%
  \addtocounter{footnote}{-1}%
  \endgroup
}

\usepackage{arxiv}

\usepackage{cite}
\usepackage[utf8]{inputenc} 
\usepackage[T1]{fontenc}    
\usepackage{hyperref}       
\usepackage{url}            
\usepackage{booktabs}       
\usepackage{amsfonts}       
\usepackage{nicefrac}       
\usepackage{microtype}      
\usepackage{lipsum}		
\usepackage{graphicx}
\usepackage{doi}

\usepackage{tikz}
\usepackage{textcomp}
\newcommand\copyrighttext{%
\footnotesize \textcopyright 2022 IEEE. Personal use of this material is permitted.
Permission from IEEE must be obtained for all other uses, in any current or future
media, including reprinting/republishing this material for advertising or promotional
purposes, creating new collective works, for resale or redistribution to servers or
lists, or reuse of any copyrighted component of this work in other works.

\doi{10.1109/WMCS55582.2022.9866326}}
\newcommand\copyrightnotice{%
\begin{tikzpicture}[remember picture,overlay]
\node[anchor=south,yshift=20pt] at (current page.south) {\fbox{\parbox{\dimexpr\textwidth-\fboxsep-\fboxrule\relax}{\copyrighttext}}};
\end{tikzpicture}%
}

\title{A Vision Transformer Approach for Efficient Near-Field Irregular SAR Super-Resolution}

\author{ \href{https://orcid.org/0000-0002-3388-4805}{\includegraphics[scale=0.06]{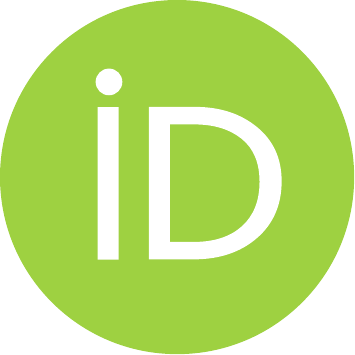}\hspace{1mm}Josiah W. Smith} \\
	Department of Electrical and Computer Engineering\\
	The University of Texas at Dallas\\
	Richardson, TX 75080 \\
	\texttt{josiah.smith@utdallas.edu} \\
	\And
	\href{https://orcid.org/0000-0001-7229-1765}{\includegraphics[scale=0.06]{orcid.pdf}\hspace{1mm}Yusef Alimam} \\
	Department of Electrical and Computer Engineering\\
	The University of Texas at Dallas\\
	Richardson, TX 75080 \\
	\And
	\href{https://orcid.org/0000-0001-7229-1765}{\includegraphics[scale=0.06]{orcid.pdf}\hspace{1mm}Geetika Vedula} \\
	Department of Electrical and Computer Engineering\\
	The University of Texas at Dallas\\
	Richardson, TX 75080 \\
	\And
	\href{https://orcid.org/0000-0001-7229-1765}{\includegraphics[scale=0.06]{orcid.pdf}\hspace{1mm}Murat Torlak}\thanks{The work of Murat Torlak (while serving at NSF) was supported by NSF.} \\
	Department of Electrical and Computer Engineering\\
	The University of Texas at Dallas\\
	Richardson, TX 75080 \\
	\texttt{torlak@utdallas.edu} \\
}

\date{}

\renewcommand{\headeright}{\textit{Proc. IEEE WMCS} (Accepted)}
\renewcommand{\undertitle}{\textit{Proc. IEEE WMCS} (Accepted)}
\renewcommand{\shorttitle}{A Vision Transformer Approach}

\hypersetup{
pdftitle={A Vision Transformer Approach for Efficient Near-Field Irregular SAR Super-Resolution},
pdfsubject={cs.CV, cs.AI, eess.SP},
pdfauthor={Josiah W.~Smith, Yusef Alimam, Geetika Vedula, Murat Torlak},
pdfkeywords={(SAR)},
}

\begin{document}
\maketitle
\copyrightnotice

\begin{abstract}
In this paper, we develop a novel super-resolution algorithm for near-field synthetic-aperture radar (SAR) under irregular scanning geometries. 
As fifth-generation (5G) millimeter-wave (mmWave) devices are becoming increasingly affordable and available, high-resolution SAR imaging is feasible for end-user applications and non-laboratory environments. 
Emerging applications such freehand imaging, wherein a handheld radar is scanned throughout space by a user, unmanned aerial vehicle (UAV) imaging, and automotive SAR face several unique challenges for high-resolution imaging.
First, recovering a SAR image requires knowledge of the array positions throughout the scan. 
While recent work has introduced camera-based positioning systems capable of adequately estimating the position, recovering the algorithm efficiently is a requirement to enable edge and Internet of Things (IoT) technologies.
Efficient algorithms for non-cooperative near-field SAR sampling have been explored in recent work, but suffer image defocusing under position estimation error and can only produce medium-fidelity images.
In this paper, we introduce a mobile-friend vision transformer (ViT) architecture to address position estimation error and perform SAR image super-resolution (SR) under irregular sampling geometries.
The proposed algorithm, Mobile-SRViT, is the first to employ a ViT approach for SAR image enhancement and is validated in simulation and via empirical studies.
\end{abstract}

\blfootnote{This work was supported in part by Texas Instruments through the Foundational Technology Research Centre and the Texas Analog Center of Excellence.}


\section{Introduction}
Millimeter-wave (mmWave) imaging systems have garnered significant attention in recent years as ultrawideband (UWB) devices are becoming increasingly affordable.
Radar devices operating in the mmWave spectrum have been applied in applications such as concealed weapon detection \cite{sheen2001three}, non-destructive testing \cite{ghasr201330}, automotive imaging \cite{garcia20203DSARProcessing}, and hand gesture recognition and tracking \cite{smith2021sterile,zheng2021dynamic,smith2021An}. 
Synthetic aperture radar (SAR) imaging is of particular interest and involves scanning a radar across space to create a synthetic aperture much larger than the radar itself \cite{yanik2019sparse}. 
Traditional SAR imaging requires high-precision laboratory equipment for exact positioning of the antennas throughout the scan.
Efficient SAR imaging algorithms have been explored extensively in the literature \cite{sheen2001three,yanik2020development,smith2020nearfieldisar} leveraging the fast Fourier transform (FFT) to recover the image from the radar data. 
These efficient algorithms strictly require specific synthetic aperture geometries, e.g., planar \cite{sheen2001three,yanik2020development}, cylindrical \cite{smith2020nearfieldisar}, etc., to achieve high-resolution imaging. 
However, with the emergence of fifth-generation (5G) and Internet of Things (IoT) technologies, near-field SAR sensing at the edge has already received attention at both the system and algorithm levels \cite{alvarez2021towards,smith2022efficient}.
One application of interest is known as freehand imaging and involves using a smartphone or handheld radar device to perform the SAR scan. 
Although these applications operate in similar frequencies to traditional laboratory SAR \cite{yanik2019cascaded}, they suffer from two primary constraints: 1) the resulting SAR array generally does not conform to the traditional geometries required by existing algorithms \cite{sheen2001three,yanik2020development,smith2020nearfieldisar} and 2) since the image computation typically takes place on a low-power or mobile device, the computational load must be reduced compared to conventional imaging. 
As a result, recovering high-fidelity images under such conditions remains an open challenge.

Previous work on efficient near-field SAR imaging proposes a decomposition of the irregular multiple-input multiple-output SAR (MIMO-SAR) array into a multi-planar imaging scenario wherein the multistatic samples are taken across a volume in space and then projected onto a reference plane for efficient image recovery, referred to as the efficient multi-planar multistatic (EMPM) algorithm \cite{smith2022efficient}.
This algorithm achieves similar imaging quality to the gold-standard backprojection algorithm (BPA) at a fraction of the computational load. 
However, prior analyses \cite{smith2022efficient,alvarez2021freehandsystem} do not take into account errors in the position estimation present in practical implementations \cite{alvarez2021towards}. 
Such position errors cause defocusing and distortion to SAR images as the algorithm improperly computes the matched filter weights based on the noisy position estimates. 
Without knowledge of the exact positions, removing distortion present in images recovered from either the EMPM or BPA remains an open challenge.
For a practical system, the EMPM is capable of efficiently reconstructing only a medium-fidelity image. 

Separately, deep learning approaches for optical image super-resolution have been extended into the radar domain for SAR image super-resolution \cite{smith2021An,gao2018enhanced,jing2022enhanced,dai2021imaging}.
Using convolutional neural network (CNN) architectures, previous efforts have seen success in improving SAR resolution \cite{gao2018enhanced,jing2022enhanced} and removing multistatic artifacts \cite{dai2021imaging}.
However, these techniques operate on SAR images collected using traditional techniques in laboratory environments and are not suitable for the irregular sampling geometry explored in this paper.
Nevertheless, deep learning has seen tremendous success in both the optical domain, for image restoration \cite{liang2021swinir} and super-resolution \cite{lim2017enhanced}, and radar domain \cite{gao2018enhanced,jing2022enhanced,dai2021imaging}.
Hence, deep learning may be a suitable solution for near-field irregular SAR artifact mitigation and super-resolution. 

Separately, recent advances in computer vision have seen a shift from CNN-based architectures towards the attention mechanism \cite{vaswani2017attention} using Vision Transformer (ViT) techniques \cite{dosovitskiy2020image_ViT} to achieve performance gains with smaller model sizes \cite{liu2021swin,sandler2018mobilenetv2,mehta2021mobilevit}. 
In \cite{mehta2021mobilevit}, the MobileViT architecture is presented leveraging a transformer architecture for image classification. 
Later, the transformer architecture was employed for optical image super-resolution and artifact mitigation in \cite{liang2021swinir}. 
Transformer techniques have appeared in recent work on radar image classification \cite{dong2021exploring} and gesture recognition \cite{zheng2021dynamic}; however, transformers have yet to be employed for SAR image super-resolution. 
In this paper, we introduce a novel transformer-based architecture for SAR image super-resolution under irregular sampling geometries called Mobile-SRViT. 
The proposed algorithm operates on images recovered by the EMPM algorithm \cite{smith2022efficient} and produces high-fidelity images of intricate targets. 
We validate our mobile-friendly algorithm using simulation and empirical data from a near-field SAR scenario with irregular scanning geometry. 

The remainder of this paper is formatted as follows.
Section \ref{sec:hffh_vit_signal_model} overviews the requisite signal model for near-field irregularly sampled SAR.
In Section \ref{sec:hffh_vit_methods}, we detail our proposed algorithm.
Experimental results are included in Section \ref{sec:hffh_vit_results} followed finally by conclusions.

\section{Signal Model}
\label{sec:hffh_vit_signal_model}
In this section, the signal model for non-cooperative SAR in the near-field is briefly introduced.
Considering a multi-planar multistatic array with transmitter (Tx) and receiver (Rx) antennas located at $(x_T,y_T,z_\ell)$ and $(x_R,y_R,z_\ell)$, respectively, the received signal at the $\ell$-th Tx/Rx pair can be modeled as
\begin{equation}
    \label{eq:received}
    s(x_T,x_R,y_T,y_R,z_\ell,t) = \iint \frac{o(x,y)}{R_\ell^T R_\ell^R} p \left( t - \frac{R_\ell^T}{c} - \frac{R_\ell^R}{c} \right) dx dy,
\end{equation}
assuming the Born approximation and an isotropic antenna, where $o(x,y)$ is known as the target reflectivity at the plane $z = z_0$, $p(t)$ is the signal at the transmitter, $t$ is the fast-time variable, $c$ is the speed of light, and $R_\ell^T$, $R_\ell^R$ are given by
\begin{align}
\label{eq:Rl}
    \begin{split}
        R_\ell^T &= \left[(x_T - x)^2 + (y_T - y)^2 + (z_\ell - z_0)^2 \right]^{\frac{1}{2}}, \\
        R_\ell^R &= \left[(x_R - x)^2 + (y_R - y)^2 + (z_\ell - z_0)^2 \right]^{\frac{1}{2}}.
    \end{split}
\end{align}
Taking the Fourier transform of (\ref{eq:received}) with respect to time yields the frequency spectrum, which can be expressed as
\begin{equation}
    \label{eq:received_k}
    s(x_T,x_R,y_T,y_R,z_\ell,f) = P(f) \iint \frac{o(x,y)}{R_\ell^T R_\ell^R} e^{-jk(R_\ell^T + R_\ell^R)} dx dy,
\end{equation}
where $P(f)$ is the spectrum of $p(t)$ and the instantaneous wavenumber is given by $k = 2 \pi f /c$.

By the analysis in \cite{smith2022efficient}, the multi-planar multistatic data can be projected to an equivalent virtual planar sampled array by
\begin{equation}
    \label{eq:multiplanar_compensation_hffh_vit}
    \hat{s}(x',y',f) \approx s(x_T,x_R,y_T,y_R,z_\ell,f) e^{j k\beta_\ell},
\end{equation}
where
\begin{equation}
    \beta_\ell = 2 d_\ell^z + \frac{(d_\ell^x)^2 + (d_\ell^y)^2}{4 Z_0},
\end{equation}
and $\hat{s}(x',y',f)$ is the virtual planar monostatic array with virtual coplanar antenna positions on the $Z_0$ plane located at the midpoint of each Tx/Rx pair. 
After this compensation step, the image can be recovered using the efficient Fourier-based range migration algorithm (RMA) yielding the 2-D reflectivity image $\hat{o}(x,y)$ \cite{yanik2019sparse,smith2022efficient}. 
Besides the EMPM, whose complexity is $O(N^2 \log(N))$, the BPA also does not require specific sampling geometry.
However, the complexity of the BPA is $O(N^4)$, which is generally prohibitive for end-user applications \cite{smith2022efficient,yanik2020development}.
Furthermore, the recovered image from either algorithm is degraded by distortion and defocusing due to errors in the position estimates. 
In attempt to address these artifacts and improve SAR imaging  resolution, we propose a novel Mobile-SRViT architecture.

\begin{figure}[h]
    \centering
    \includegraphics[width=0.3\textwidth]{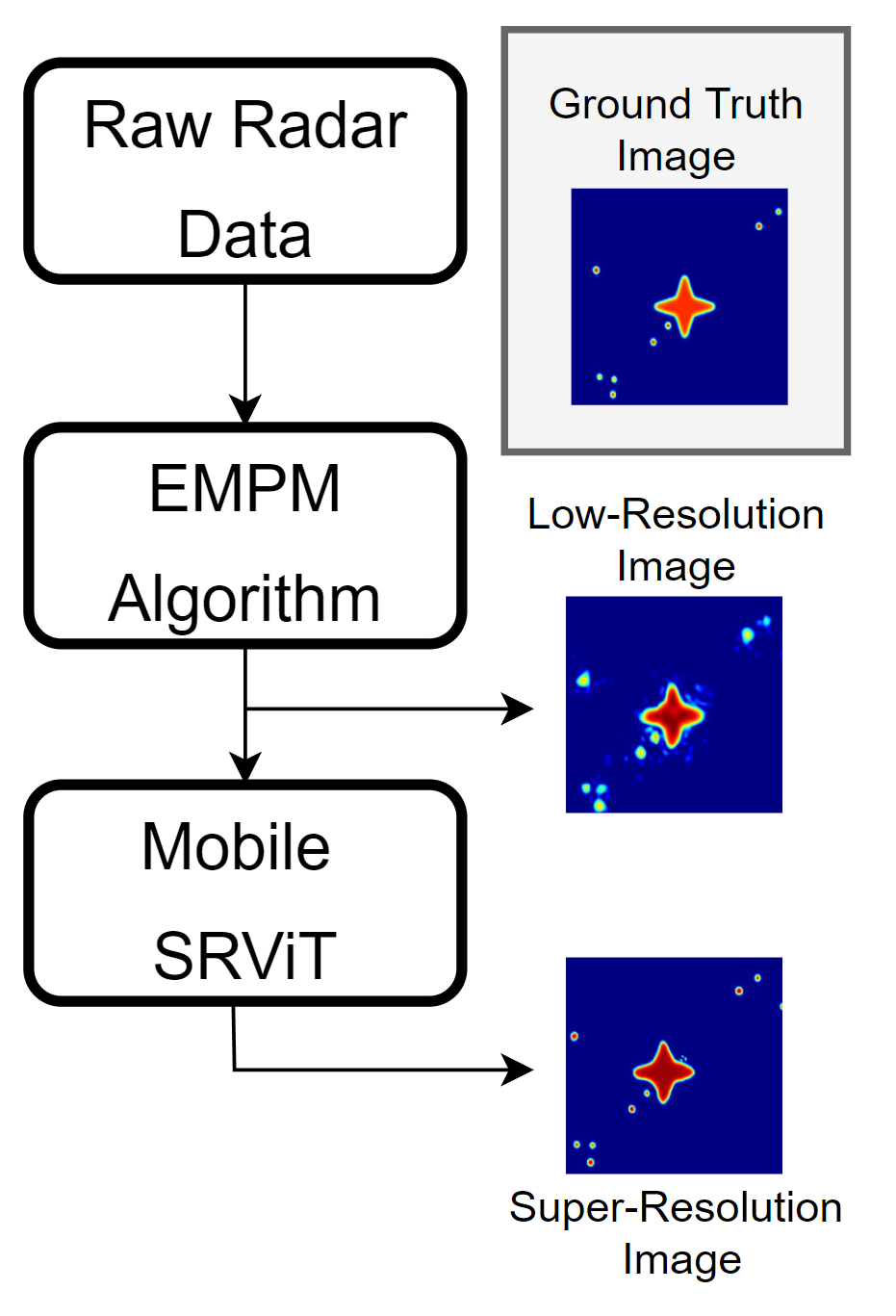}
    \caption{Operation of the Mobile-SRViT: the low resolution image produced by the EMPM algorithm is restored by the Mobile-SRViT algorithm. The ground truth image is shown for reference.}
    \label{fig:hffh_vit_flow}
\end{figure} 

\begin{figure*}[ht]
    \centering
    \includegraphics[width=\textwidth]{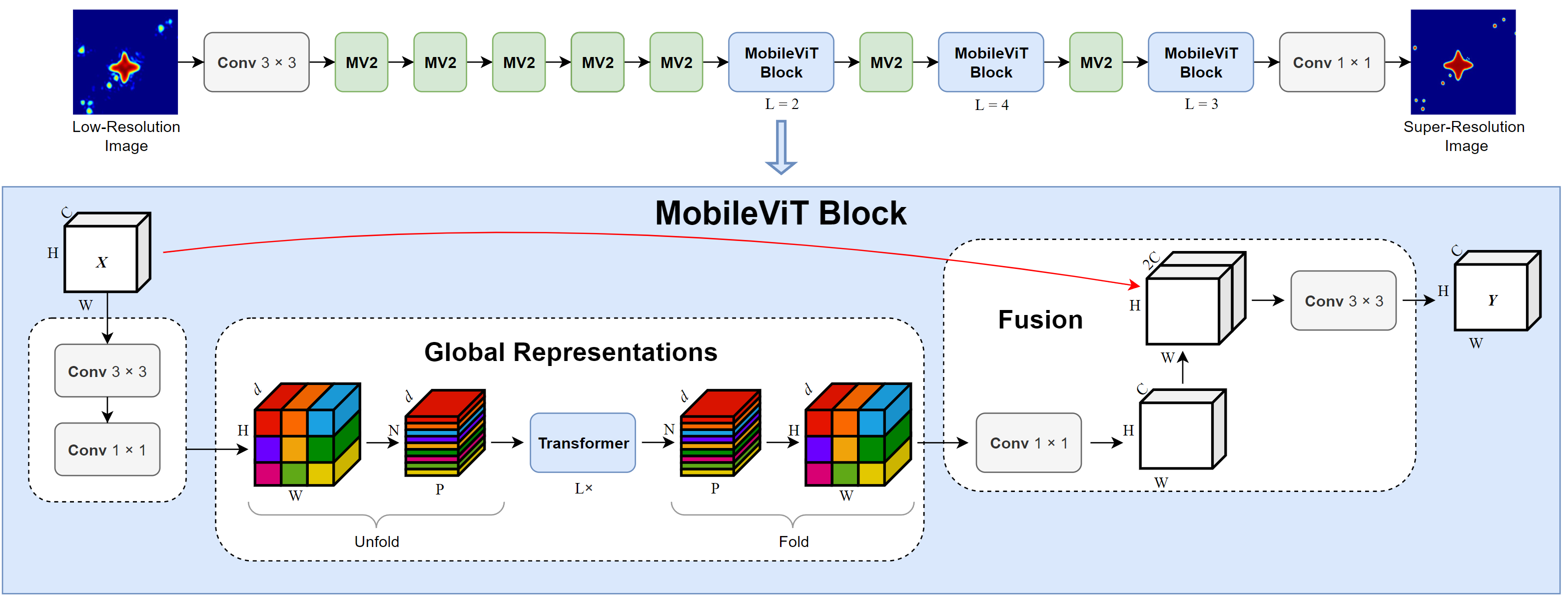}
    \caption{Mobile-SRViT architecture.}
    \label{fig:hffh_vit_net}
\end{figure*} 

\section{Irregular SAR Super-Resolution using Vision Transformers}
\label{sec:hffh_vit_methods}
In this section, we detail the proposed transformer-based approach for near-field irregular SAR super-resolution and artifact mitigation. 
An overview of the Mobile-SRViT algorithm is given in Fig. \ref{fig:hffh_vit_flow}.
The raw radar data are processed by the EMPM algorithm to produce a low-resolution image with distortion, blur, and defocusing caused by imaging non-idealities.
The proposed Mobile-SRViT operates on this image to produce a super-resolution image, restoring image quality while preserving key high-frequency details of the targets.

The architecture of the Mobile-SRViT is based on the MobileViT network employed for image classification in \cite{mehta2021mobilevit}. 
Since our network is designed for image-to-image processing, the convolution layers are modified to adhere to a fully convolutional neural network (FCNN) framework, similar to that of \cite{smith2021An}.
Fig. \ref{fig:hffh_vit_net} details the implementation of the Mobile-SRViT algorithm, where ``MV2'' refers to the MobileNetV2 block proposed in \cite{sandler2018mobilenetv2}.
Our algorithm adopts the approach of \cite{mehta2021mobilevit} such that the image is first processed by several MobileNetV2 convolution blocks before alternating between MobileViT and MobileNetV2 operations. 
The MobileViT block is intended to model the global and local information of the input data with fewer parameters than the traditional ViT \cite{dosovitskiy2020image_ViT}. 
Given an input tensor $\mathbf{X} \in \mathbb{R}^{H \times W \times C}$, where $H$ is the height, $W$ is the width, and $C$ is the number of channels, the MobileViT block first applies a $3 \times 3$ convolution layer followed by a $1 \times 1$, or pointwise, convolution layer to produce a tensor $\mathbf{X}_L \in \mathbb{R}^{H \times W \times d}$. 
The last $1 \times 1$ convolution layer reduces the channels to match the input image.
In order to learn global representations, $\mathbf{X}_L$ is unfolded into $N$ non-overlapping patches $\mathbf{X}_U \in \mathbb{R}^{P \times N \times d}$, where $P = 4$ and $N = HW/4$.
Each of the $P$ patches are processed using a transformer architecture to encode the inter-patch relationships yielding 
\begin{equation}
    \label{eq:transformer_mobilevit}
    \mathbf{X}_G(p) = \text{Transformer}(\mathbf{X}_U(p)), \quad p \in [1, \dots, P].
\end{equation}
Whereas most ViT implementations lose the positional location of each patch \cite{liu2021swin,dosovitskiy2020image_ViT}, the MobileViT retains the patch order and the pixel order within each patch. 
As a result, $\mathbf{X}_G \in \mathbb{R}^{P \times N \times d}$ can be directly folded to obtain $\mathbf{X}_F \in \mathbb{R}^{H \times W \times d}$. 
The resulting tensor $\mathbf{X}_F$ is projected to a low $C$-dimensional space via a $1 \times 1$ convolution layer before being concatenated with $\mathbf{X}$ yielding $\mathbf{X}_O \in \mathbb{R}^{H \times W \times 2C}$.
Finally, using a $3 \times 3$ convolution, $\mathbf{X}_O$ is fused to form the output tensor $\mathbf{Y}$ of identical size to $\mathbf{X}$.
Interestingly, the receptive field of the MobileViT block is $H \times W$ because $\mathbf{X}_U(p)$ encodes local information from a $3 \times 3$ region via convolutions and each pixel in $\mathbf{X}_G(p)$ encodes global information over $P$ patches \cite{mehta2021mobilevit}. 
Our implementation maintains $C = 16$ until the last MobileViT block where $C = 32$, $d = 2C$ for each MobileViT block, and $L = \{2, 4, 3\}$. 
The images are of size $256 \times 256$ and the patch size employed is $16 \times 16$.
With this architecture, the proposed Mobile SR-ViT has 69,122 parameters.
Loss is computed using the pixel-to-pixel L1 metric as
\begin{equation}
\label{eq:L1_loss}
    L_{p2p} = ||\mathbf{X}_{SR} - \mathbf{X}_{HR}||_1.
\end{equation}
Prior attempts at near-field SAR super-resolution have been purely CNN-based \cite{smith2021An,gao2018enhanced,jing2022enhanced,dai2021imaging}, but the Mobile-SRViT detailed in this paper is the first to leverage a transformer architecture for SAR imaging. 

\begin{figure*}[ht]
\centering
    \begin{subfigure}[b]{0.245\textwidth}
         \centering
         \includegraphics[width=\textwidth]{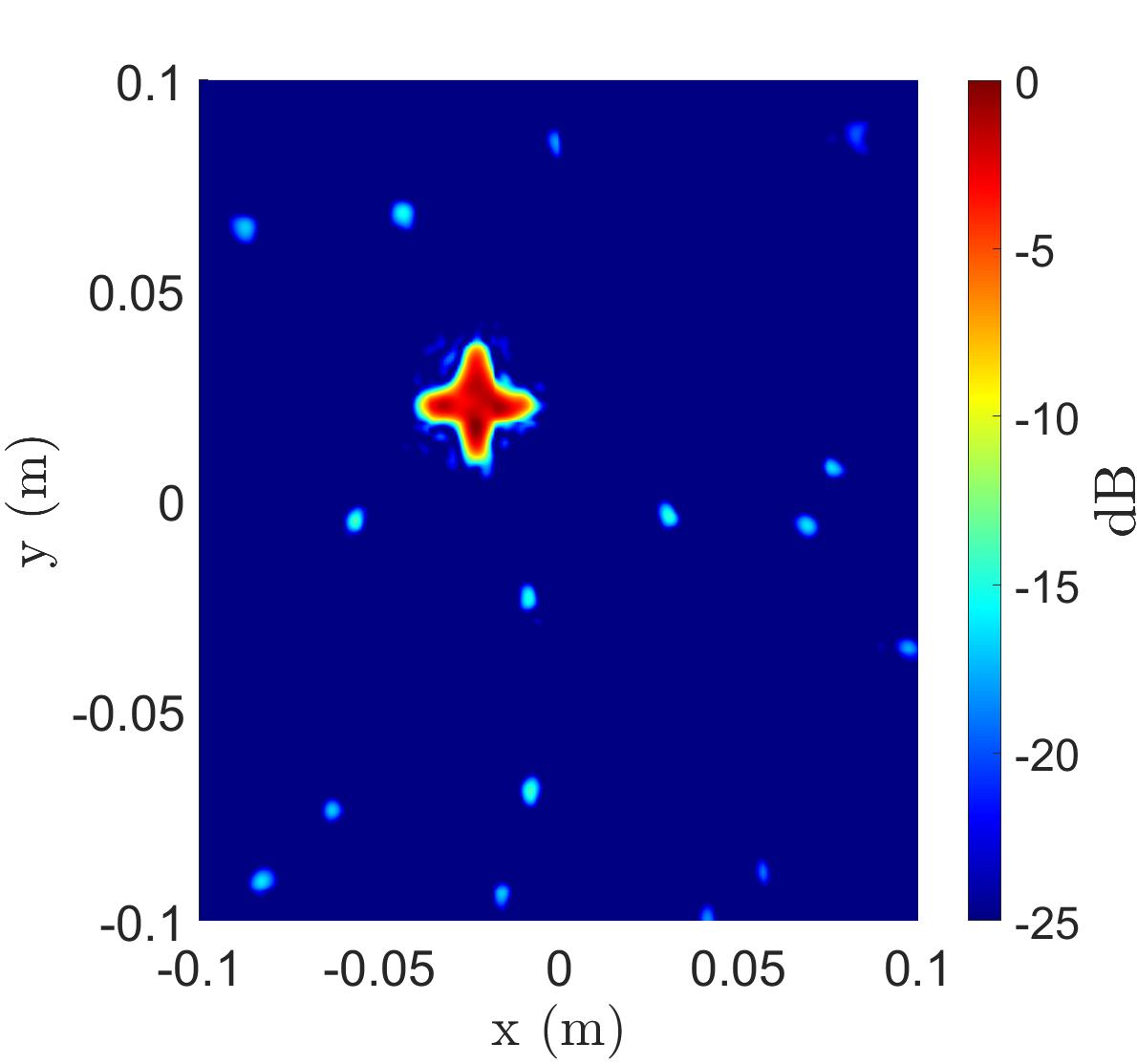} 
         \caption{}
         \label{fig:test270_lr}
    \end{subfigure}
    \begin{subfigure}[b]{0.245\textwidth}
         \centering
         \includegraphics[width=\textwidth]{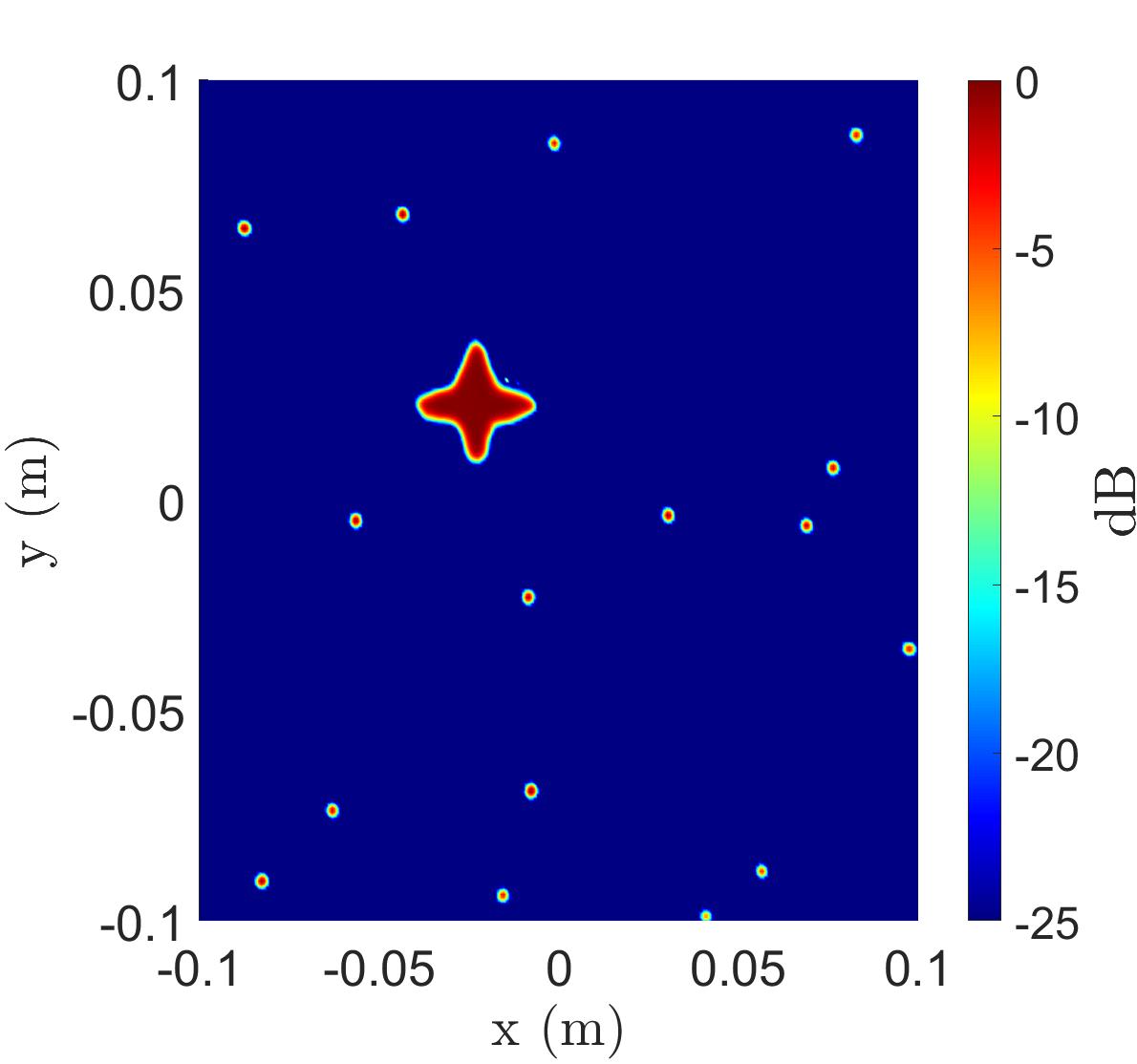} 
         \caption{}
         \label{fig:test270_sr}
    \end{subfigure}
    \begin{subfigure}[b]{0.245\textwidth}
         \centering
         \includegraphics[width=\textwidth]{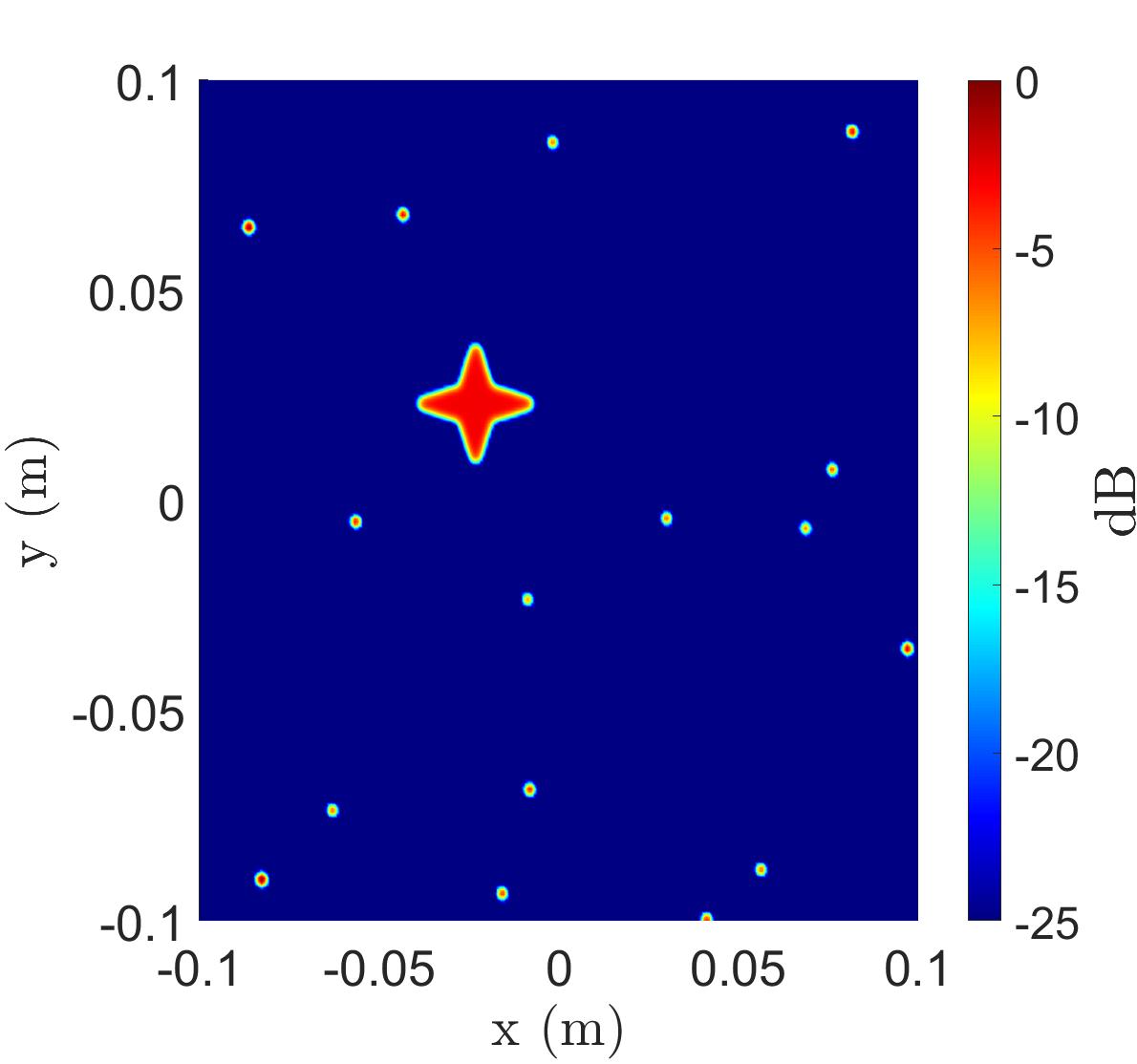} 
         \caption{}
         \label{fig:test270_hr}
    \end{subfigure}
    \vskip\baselineskip
    \begin{subfigure}[b]{0.245\textwidth}
         \centering
         \includegraphics[width=\textwidth]{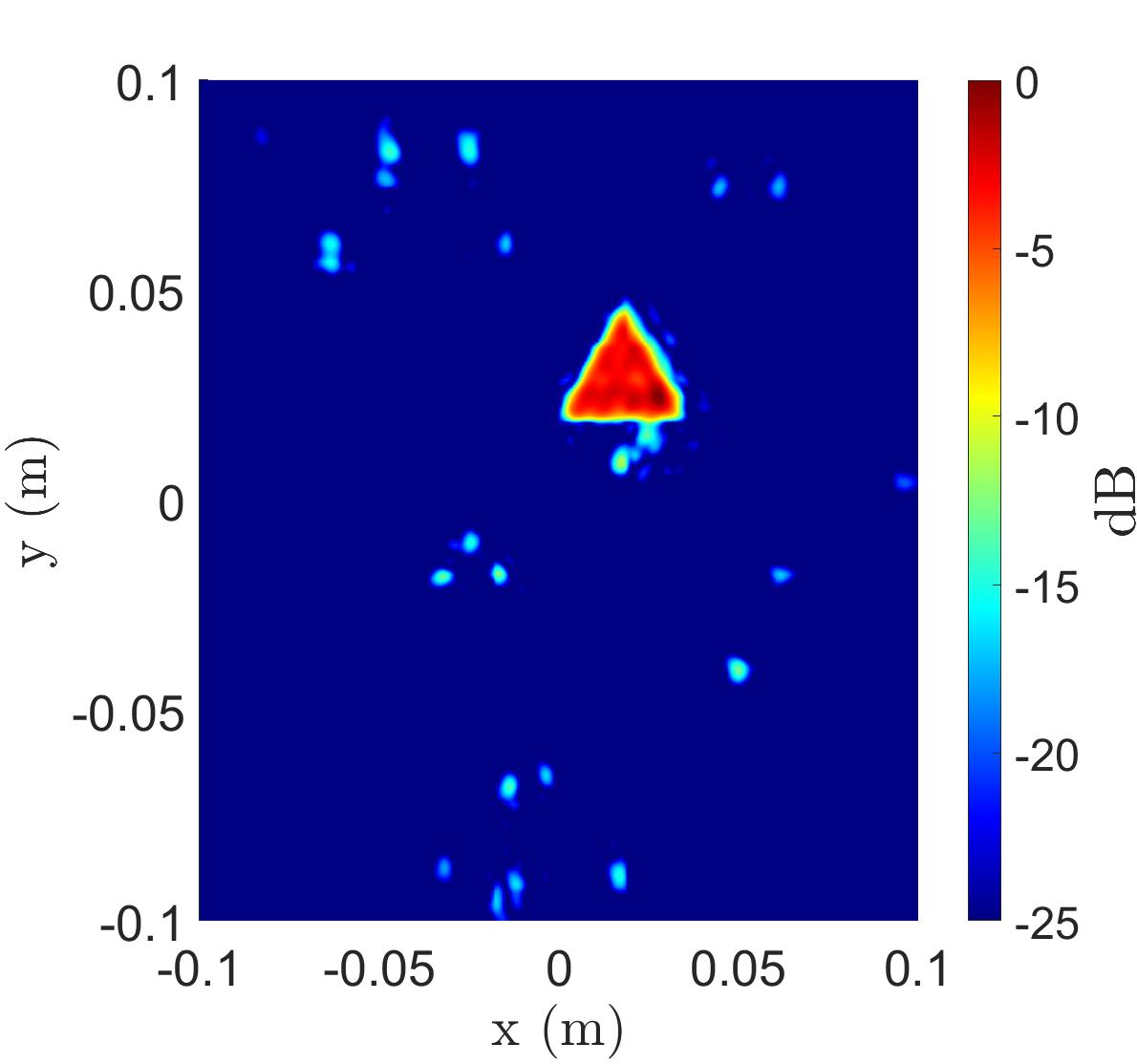} 
         \caption{}
         \label{fig:test707_lr}
    \end{subfigure}
    \begin{subfigure}[b]{0.245\textwidth}
         \centering
         \includegraphics[width=\textwidth]{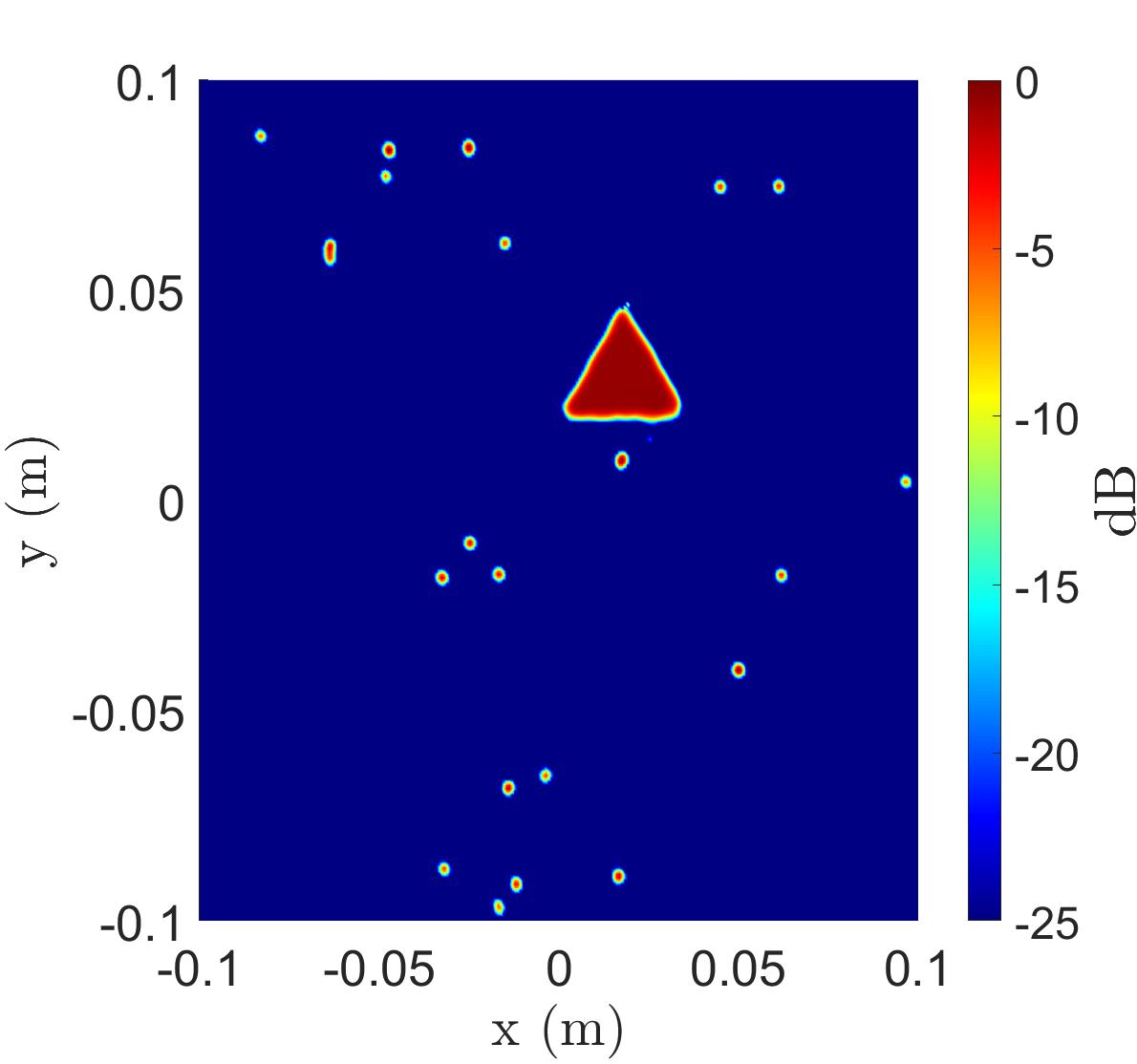} 
         \caption{}
         \label{fig:test707_sr}
    \end{subfigure}
    \begin{subfigure}[b]{0.245\textwidth}
         \centering
         \includegraphics[width=\textwidth]{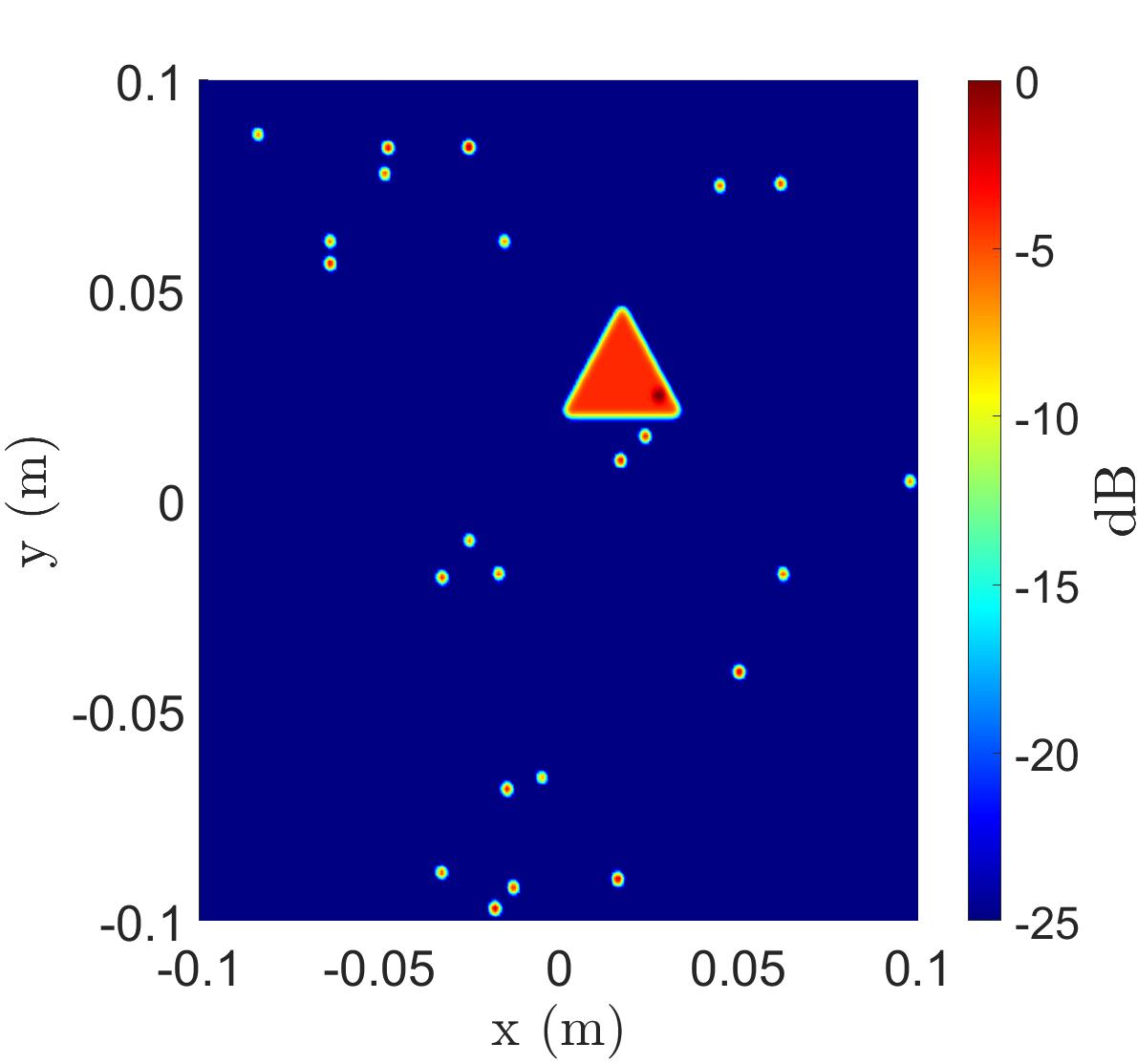} 
         \caption{}
         \label{fig:test707_hr}
    \end{subfigure}
\caption{Imaging results using the Mobile-SRViT on synthetic data. The images in the first column, (a) and (d), are produced by the EMPM algorithm and input to the Mobile-SRViT. The images in the second column, (b) and (e), are the super-resolution images output from the Mobile-SRViT. The images in the third column, (c) and (f), are the ground truth images.}
\label{fig:hffh_vit_results_sim}
\end{figure*}

\subsection{Training the Mobile-SRViT}
\label{subsec:hffh_vit_training}
Our efficient algorithm is trained for 50 epochs using an ADAM optimizer on a single RTX3090 GPU with 24 GB of memory using 4096 samples for the training process and 1024 samples for evaluation. 
The training data is generated using the procedure detailed in \cite{smith2021An,smith2022efficient,gao2018enhanced}.
Whereas prior methods \cite{smith2021An,gao2018enhanced,jing2022enhanced,dai2021imaging} train on SAR images from exclusively point scatterers, we introduce more sophisticated targets consisting of solid and hollow objects in addition to randomly placed point scatterers, such as the example images in Fig. \ref{fig:hffh_vit_flow}.
By including more complex targets in the training dataset, our algorithm is able to generalize for solid and hollow targets. 
Each sample is generated with additive white Gaussian noise (AWGN) with a signal-to-noise ratio in the range $[-10, 50]$ dB and includes AWGN positioning errors with a standard deviation of 1 mm along the $x$, $y$, and $z$-directions to emulate a practical scenario. 
The training process was approximately 6 hours with an inference time during validation of $10$ ms per sample. 

\section{Experimental Results}
\label{sec:hffh_vit_results}
In this section, we conduct simulation and empirical experiments to verify the proposed algorithm.
To evaluate the performance of the SRViT algorithm, we first compare the numerical performance of the Mobile-SRViT to the EMPM, BPA, and RMA. 
The RMA requires a planar sampling and is unable to recover an image from an irregularly sampled array, as discussed in \cite{smith2022efficient}. 
The gold-standard BPA has no requirements for SAR array geometry and is well-suited for irregular scanning geometries.
However, it is computationally prohibitive, particularly for mobile applications, as it computes the pixel-wise matched filter for every sampling location and frequency.
The EMPM, on the other hand, is an efficient RMA-based algorithm but assumes the samples are taken across a small volume relative to a reference plane $Z_0$ \cite{smith2022efficient}. 
The proposed Mobile-SRViT algorithm attempts to compensate for the distortion present in the EMPM images due to these assumptions in addition to imaging nonidealities. 

\begin{table}[h]
\caption{Quantitative performance of the Mobile-SRViT compared to the BPA, EMPM, and RMA.} 
\centering
  \begin{tabular}{ c||c|c|c|c } 
    Metrics & Mobile-SRViT & {BPA} & {EMPM} & {RMA} \\
    \hline \hline
    PSNR (dB) & $\mathbf{36.907}$ & $26.33$ & $20.20$ & $10.158$\\ 
    \hline
    RMSE & $\mathbf{0.015}$ & $0.044$ & $0.105$ & $0.276$\\
    \hline
    Time (s) & $1.113$ & $1324.8$ & $\mathbf{1.103}$ & $\mathbf{1.103}$\\
    \hline \hline
  \end{tabular}
\label{table:hffh_vit_performance}
\end{table}

Using a test dataset consisting of 1024 samples similar to those in the training dataset but never seen by the network, we apply the Mobile-SRViT, BPA, EMPM, and RMA to the samples and measure the peak signal-to-noise ratio (PSNR), root-mean-square error (RMSE), and computation time per sample.
Results are shown in Table \ref{table:hffh_vit_performance} with the best evaluations marked in bold-face.
All experiments are conducted on a desktop PC equipped with a 12-core AMD Ryzen 9 3900X running at 4.6 GHz with 64 GB of memory.
As expected, the RMA is unable to achieve image quality, in terms of PSNR and RMSE, comparable to the BPA or EMPM but is highly efficient.
The EMPM achieves identical computation time to the RMA with much higher PSNR and lower RMSE, approaching the BPA.
On the other hand, the BPA boasts the highest PSNR and lowest RMSE of the classical algorithms but requires a significantly large amount of computation time.
The Mobile-SRViT is superior to the other algorithms, even outperforming the BPA in PSNR and RMSE, with a total computation time of 1.113 seconds required to compute the EMPM and pass the image through the network. 
This qualitative analysis demonstrates the superiority of the proposed method in comparison to previous techniques in terms of both computational efficiency and image quality.

We further validate the performance of the proposed algorithm via visual inspection of both simulated and empirical data.
Two samples from the testing dataset are compared in Fig. \ref{fig:hffh_vit_results_sim}.
For each sample, the proposed SAR super-resolution network is able to recover the solid object in addition to the point scatterers and mitigate distortion caused by position estimation errors and system limitations.
The super-resolution images, Figs. \ref{fig:test270_sr} and \ref{fig:test707_sr}, are quite similar to the ideal images, Figs. \ref{fig:test270_hr} and \ref{fig:test707_hr}, showing an improvement over the medium-fidelity images recovered by the EMPM.

\begin{figure}[h]
\centering
    \begin{subfigure}[b]{0.45\textwidth}
         \centering
         \includegraphics[width=\textwidth]{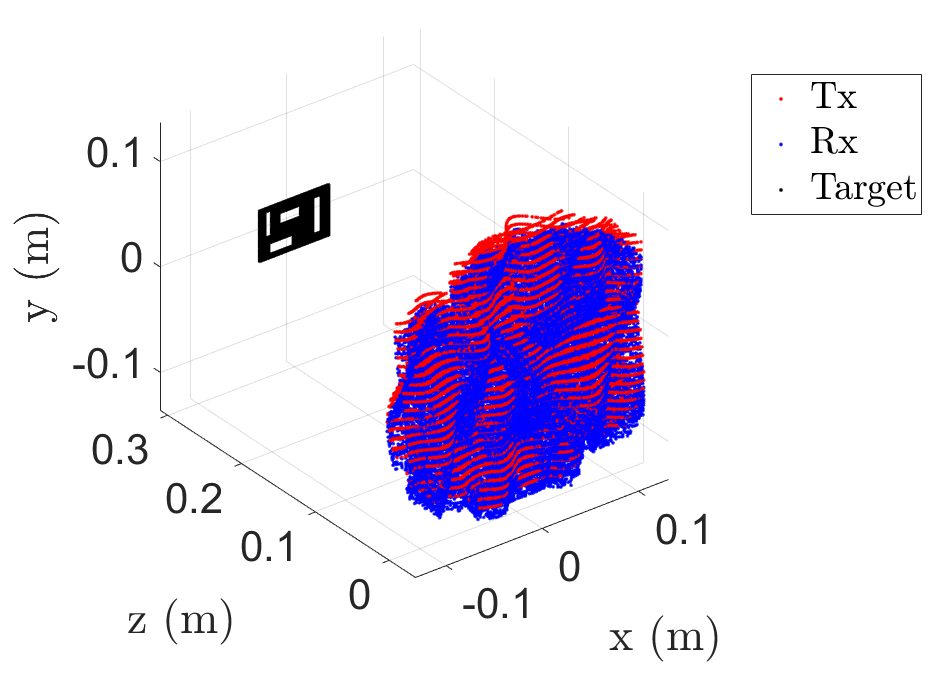}
         \caption{}
         \label{fig:irregular_geometry_example}
    \end{subfigure}
    \vskip\baselineskip
    \begin{subfigure}[b]{0.35\textwidth}
         \centering
         \includegraphics[width=\textwidth]{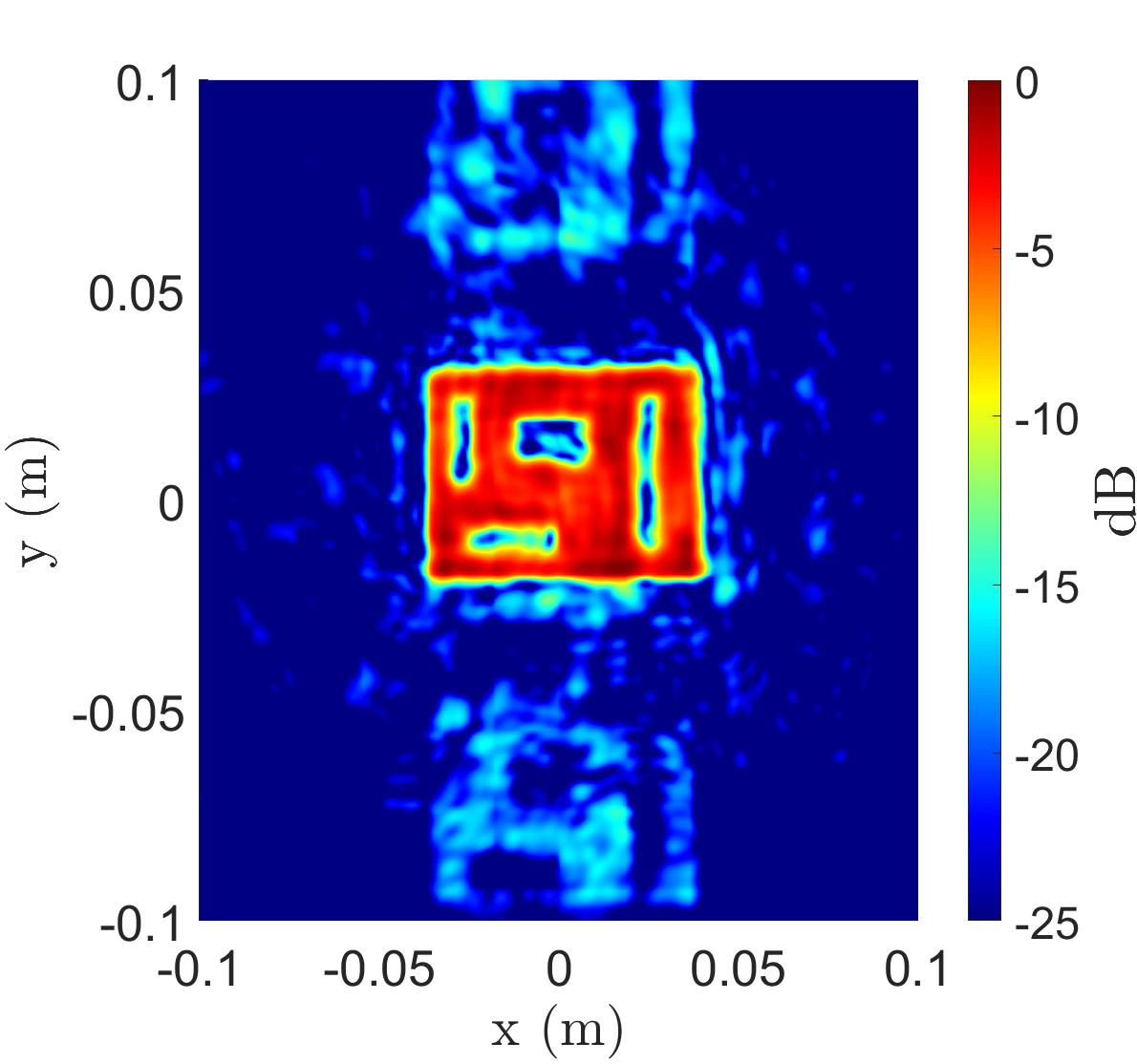}
         \caption{}
         \label{fig:exp1_lr}
    \end{subfigure}
    \begin{subfigure}[b]{0.35\textwidth}
         \centering
         \includegraphics[width=\textwidth]{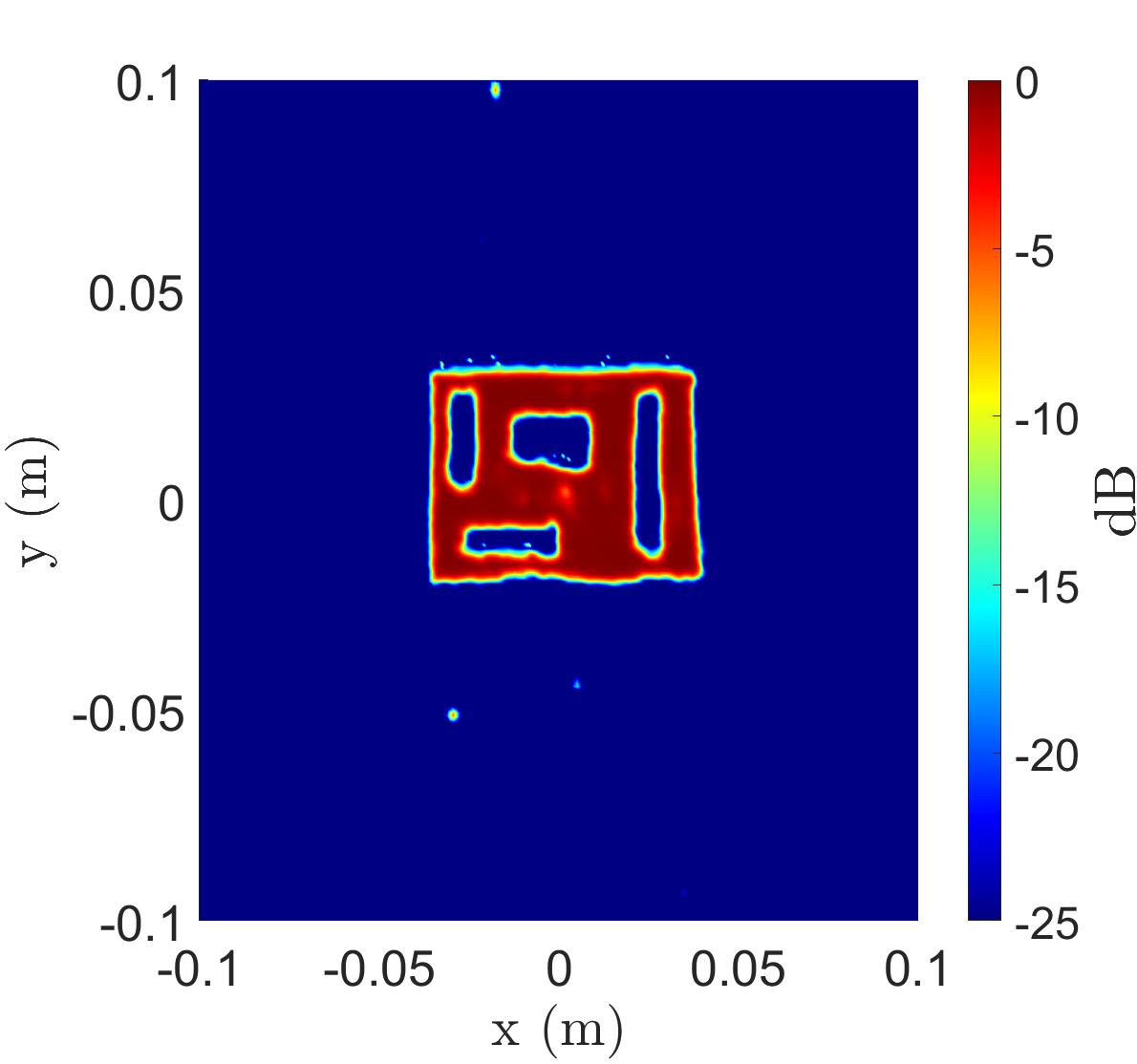}
         \caption{}
         \label{fig:exp1_sr}
    \end{subfigure}
\caption{Imaging results using the Mobile-SRViT on empirical data from a near-field SAR scenario with irregular sampling geometry and positioning error: (a) SAR sampling geometry, (b) the image reconstructed by the EMPM and (c) the super-resolution image produced by the Mobile-SRViT.}
\label{fig:hffh_vit_results_sim_real}
\end{figure}

To evaluate the performance of our proposed algorithm on empirical data, we first perform a SAR scan with irregular scanning geometry, as shown in Fig. \ref{fig:irregular_geometry_example}.
After reconstructing the image with the EMPM, shown in Fig. \ref{fig:exp1_lr}, the Mobile-SRViT is applied to achieve the super-resolution image shown in \ref{fig:exp1_sr}.
The proposed algorithm not only recovers a better-resolved image but also mitigates multistatic artifacts visible in the EMPM image \cite{smith2022efficient,yanik2019sparse}. 

\section{Conclusion}
\label{sec:hffh_vit_conclusion}
In this paper, we introduced a novel algorithm for near-field SAR super-resolution under irregular sampling geometries using a vision transformer architecture. 
Compared to previous methods for SAR super-resolution, our improved technique addresses the more challenging problem of image enhancement under non-ideal sampling conditions. 
The proposed algorithm enables numerous applications such as freehand smartphone imaging, UAV SAR, and automotive imaging. 
Using the efficient medium-fidelity EMPM algorithm developed in \cite{smith2022efficient}, we train a novel image-to-image network using state-of-the-art CNN \cite{sandler2018mobilenetv2} and ViT \cite{mehta2021mobilevit} techniques suitable for mobile applications with low latency and a small model size. 
The robust algorithm is verified in simulation and empirical studies to outperform the state-of-the-art techniques in terms of both image quality and computational complexity. 

\bibliography{mega_bib}
\bibliographystyle{IEEEtran}

\end{document}